\pdfoutput=1

\documentclass[11pt]{article}

\usepackage[final]{acl}
\usepackage[breakable]{tcolorbox}

\usepackage{times}
\usepackage{latexsym}
\usepackage{multirow}

\usepackage[T1]{fontenc}

\usepackage[utf8]{inputenc}

\usepackage{microtype}

\usepackage{inconsolata}
\usepackage{amsmath} 
\usepackage{graphicx}
\usepackage{booktabs}
\usepackage{times}
\usepackage{latexsym}
\usepackage{multirow}
\usepackage{xcolor}
\usepackage{colortbl}
\usepackage{multirow}
\usepackage{array}

\newcommand{\posval}[1]{\cellcolor[HTML]{e7f0e5}#1}
\newcommand{\highposval}[1]{\cellcolor[HTML]{bfd8b6}#1}
\newcommand{\negval}[1]{\cellcolor[HTML]{ffb6b8}#1}

\title{Detecting Errors through Ensembling Prompts (DEEP):\\An End-to-End LLM Framework for Detecting Factual Errors}

\newcommand{\alexnote}[1]{\ifdefined\showalexnotes\textcolor{blue}{#1}\else\fi}

\author{Alex Chandler \\
  University of Texas at Austin \\
  \texttt{alex.chandler@utexas.edu} \\\And
  Devesh Surve \\
  Northeastern University \\
  surve.de@northeastern.edu\\\And
  Hui Su\\
  Fidelity Investments \\  
  \texttt{Hui.Su@fmr.com} \\}

\begin{document}
\maketitle
\begin{abstract}
Accurate text summarization is one of the most common and important tasks performed by Large Language Models, where the costs of human review for an entire document may be high, but the costs of errors in summarization may be even greater. We propose
\textit{Detecting Errors through Ensembling Prompts (DEEP)} - an end-to-end large language model framework for detecting factual errors in text summarization. Our framework uses a diverse set of LLM prompts to identify factual inconsistencies, treating their outputs as binary features, which are then fed into ensembling models. We then calibrate the ensembled models to produce empirically accurate probabilities that a text is factually consistent or free of hallucination. We demonstrate that prior models for detecting factual errors in summaries perform significantly worse without optimizing the thresholds on subsets of the evaluated dataset. Our framework achieves state-of-the-art (SOTA) balanced accuracy on the AggreFact-XSUM FTSOTA, TofuEval Summary-Level, and HaluEval Summarization benchmarks in detecting factual errors within transformer-generated text summaries. It does so without any fine-tuning of the language model or reliance on thresholding techniques not available in practical settings.\footnote{Code and data are available on \href{https://github.com/achandlr/DEEP}{GitHub}.}
\end{abstract}

\section{Introduction}
The advancement of cutting-edge Large Language Models (LLMs) like  GPT-4, Claude 3, LLaMA-2, and Gemini variants introduces a significant challenge: despite producing content that is linguistically coherent, their outputs frequently contain misleading or false information, often referred to as hallucinations or factual inconsistencies. Hallucinations in Large Language Models refer to instances where the model generates usually plausible but entirely fabricated information. Factual inconsistencies, a specific type of hallucination, occur when generated text contradicts the source material or other well-established facts not explicitly mentioned in the source.

Traditional automatic evaluation methodologies like ROUGE \citep{lin-2004-rouge}, METEOR \citep{banerjee-lavie-2005-meteor}, and BLEU \citep{papineni2002bleu} have been instrumental in assessing Natural Language Generation tasks. However, numerous studies demonstrate the lack of correlation between initial automatic evaluation models and human judgment in tasks such as machine translation  \citep{callison-burch-etal-2006-evaluating, R2006SomeII}, image captioning \citep{cui2018learning}, and notably, factuality \citep{fu2023gptscore, mao2023gpteval}. In particular, these models struggle to capture semantic equivalence when there are substantial discrepancies in length, syntax, and wording between two texts \citep{guo-vosoughi-2023-length, 10.1007/978-3-540-30586-6_38}. Consequently, specialized models \citep{laban2021summac, kryściński2019evaluating, goyal2021annotating} have been developed to assess textual factual consistency, verifying the truthfulness of a claim or summary based on given ground truth textual content.

However, existing models, often fine-tuned variants of RoBERTa \citep{liu2019roberta} for assessing factual consistency, exhibit significant limitations. As highlighted in \citet{tang2023understanding}, these models show reduced effectiveness in detecting factual inconsistencies in content produced by recent state-of-the-art text-generating models. Ensemble learning is the practice of merging the outputs of multiple models to produce a more accurate prediction \citep{dietterich2000ensemble}. \citet{forbes2023metric} demonstrated that ensembling factual consistency models by calculating their weighted mean surpassed the performance of individual models in detecting hallucinations within a small dataset of GPT-3-generated Wikipedia abstractive summaries.



In light of these limitations, this study evaluates benchmarks exclusively featuring summaries from recent transformer-based language models. This approach more accurately reflects actual usage scenarios, where users commonly need to validate texts generated by newer LLMs rather than texts from older text generation models. We assess factual consistency using the AggreFact-XSUM FTSOTA, TofuEval Summary-Level, and HaluEval Summarization \citep{tang2023understanding, tang2024tofueval, li2023halueval} benchmarks, consisting of transformer-generated abstractive summaries featuring hallucinations that existing models struggle to identify.


Existing factual consistency encoder models output numerical scores, requiring thresholding to map the scores to binary labels. However, \citet{tang2023understanding} demonstrates that the optimal threshold for factual consistency models varies depending on the recentness of the summarization model within the AggreFact dataset. Previous studies that report the performance of these factual consistency models have fine-tuned each model's threshold using the development subset of the same dataset under evaluation \citep{laban2021summac, fabbri-etal-2022-qafacteval, tang2023understanding, tang2024tofueval}. 

This approach is problematic and unrealistic, as it assumes access to labeled data from the target dataset, which may not be available in real-world scenarios. In this study, we benchmark five popular state-of-the-art factual consistency models and demonstrate a substantial decline in performance when thresholds are learned from different datasets or set to their default midpoint. 

\begin{figure}[ht!]
  \centering
  \includegraphics[width=\linewidth]{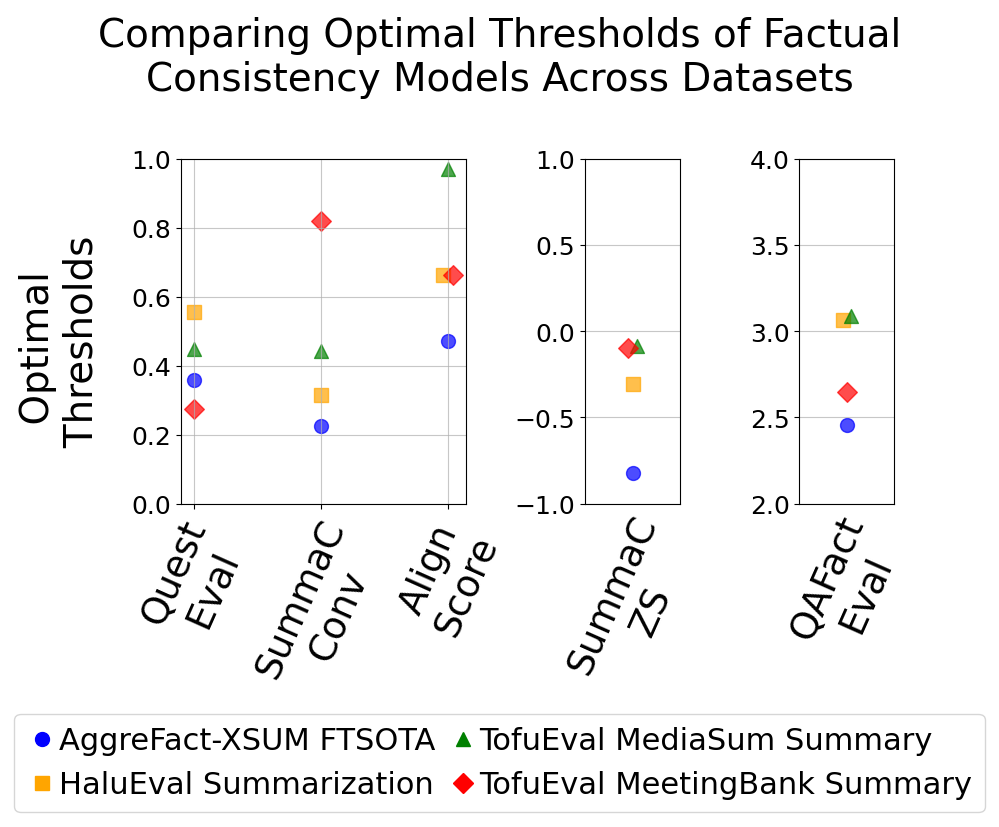}
  \caption{Optimal thresholds of factual consistency models when set to maximize balanced accuracy on each test dataset.}
  \label{fig:OptimalThresholds}
\end{figure}

\begin{figure}[ht!]
  \centering
  \includegraphics[width=\linewidth]{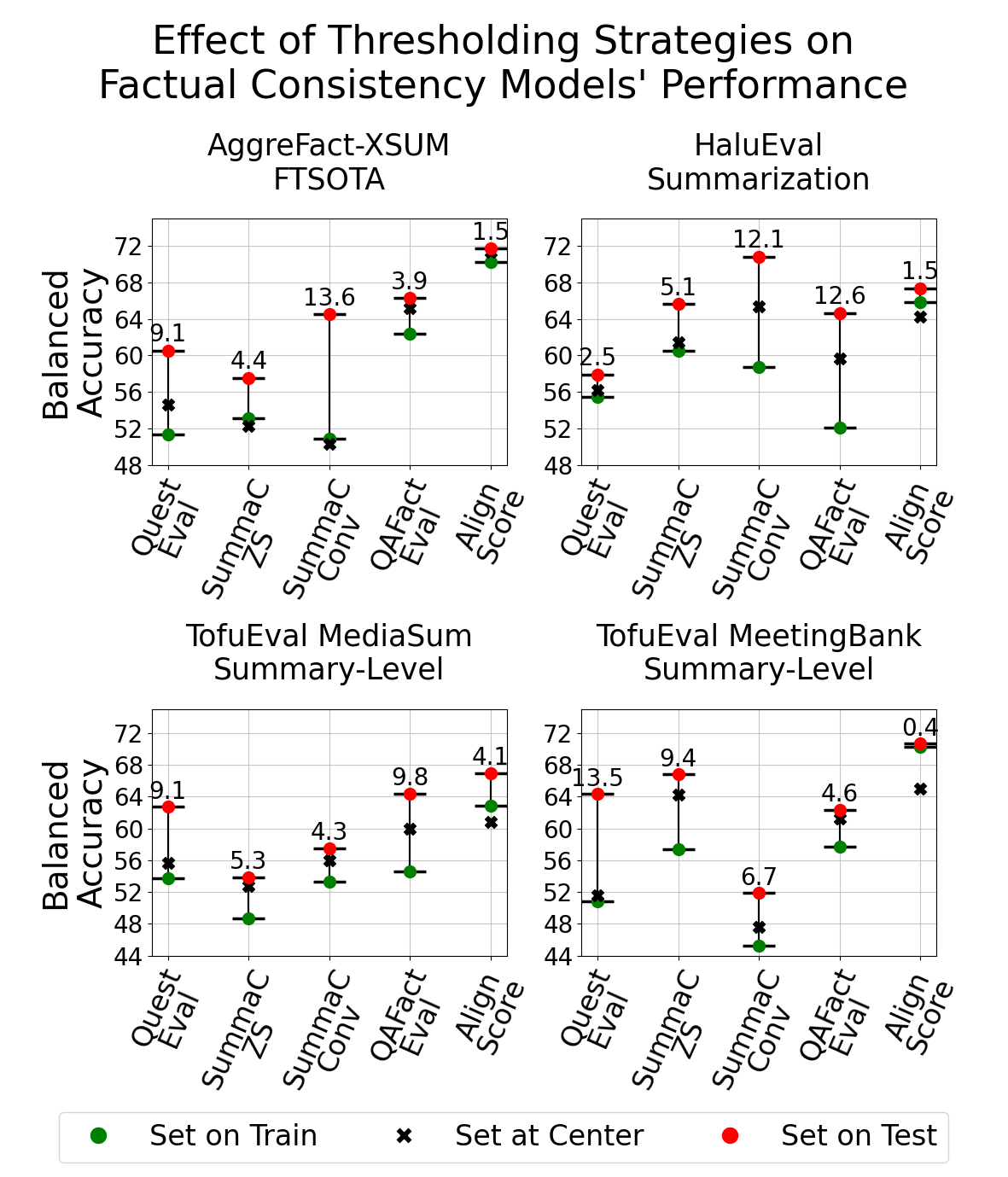}
  \caption{
  Factual consistency models' performance varies significantly based on threshold optimization strategy. The bars show balanced accuracy for factual consistency models under three threshold optimization strategies: optimizing the thresholds on the test dataset ("Optimizing on Test"), setting thresholds to the midpoint of each model's score range ("Optimizing at Center"), or optimizing on all datasets except the test set ("Optimizing on Train"), which reflects a realistic scenario of applying the model to unseen data. Numbers above bars quantify the decrease in balanced accuracy when thresholds are optimized on non-test data compared to the test dataset itself, underscoring the difficulty of effectively applying these models to unseen data in practice.
  }
  \label{fig:EncoderDifferenceInValuesVersion2Line}
\end{figure}

We extend the findings of \citet{tang2023understanding} and discover that the optimal threshold for each factual consistency model varies widely across different datasets, even when evaluating text generated solely from recent summarization models. Figure \ref{fig:OptimalThresholds} reveals that, even when considering only datasets with summaries from recent transformer models, the optimal linear threshold for maximizing balanced accuracy differs widely for each factual consistency model, covering a broad range of their possible output scores.\footnote{ QuestEval, SummaC-Conv, and AlignScore (with mode $nli\_sp$) generate scores from 0 to 1. SummaC-ZS scores range from -1 to 1. QAFactEval scores range from 0 to 5.} Moreover, Figure \ref{fig:EncoderDifferenceInValuesVersion2Line} demonstrates that optimizing thresholds on non-test data or setting them to the midpoint of each model's score range substantially reduces balanced accuracy compared to test-set optimization. The reliance of these models on dataset-specific thresholds, as demonstrated by our findings, limits their practical utility in evaluating factual consistency across a diverse range of text without further fine-tuning or adjustments.

\begin{figure*}[ht!]
  \centering
  \vspace{1em}
    \includegraphics[width=\linewidth]{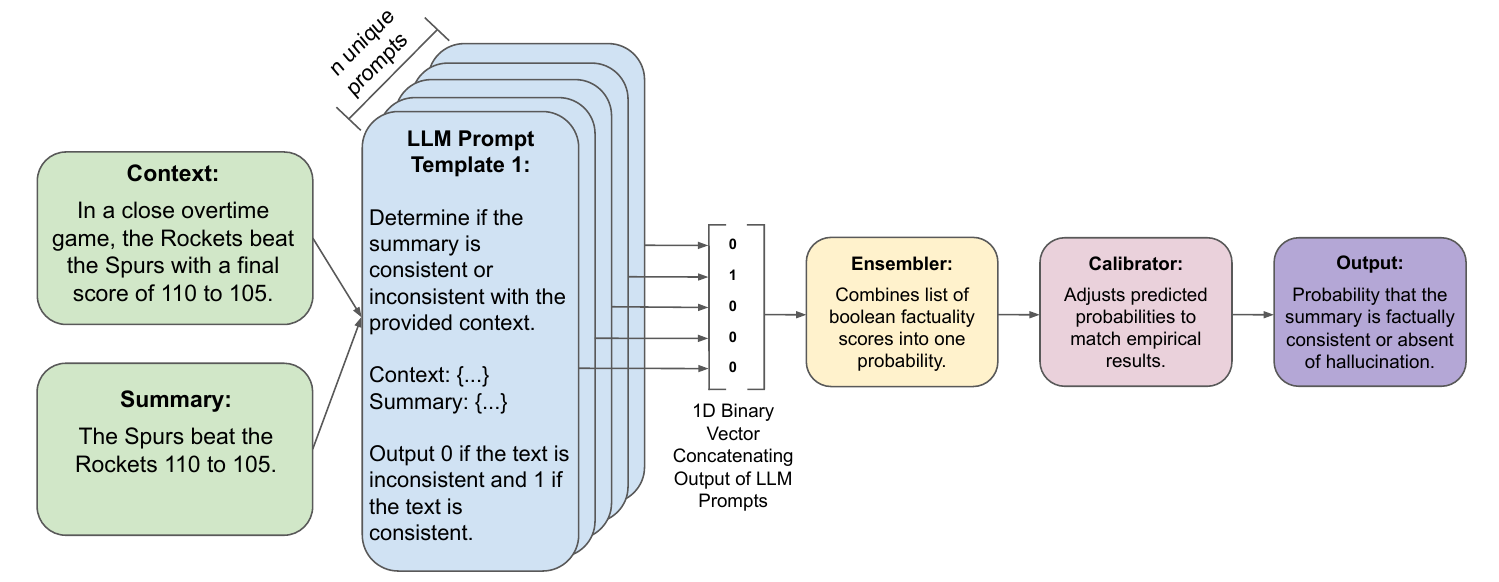}
    \caption{Diagram of our end-to-end framework.
    }
    \label{fig:endToEndPipeline}
\end{figure*}

Previous efforts to use LLMs for identifying factual inconsistencies include \citet{wang2023chatgpt}, who prompted ChatGPT to return numerical factuality scores for summarization, and \citet{luo2023chatgpt}, who used ChatGPT to produce binary factuality judgments. However, \citet{tang2023understanding} showed that these existing prompts had shown poor performance on detecting factual consistencies over the AggreFact FTSOTA benchmark of transformer-generated summaries. 
. However, reliably mapping the generated token probabilities from the reasoning steps to the final answer's confidence remains unclear, especially considering factors like temperature and sampling methods that influence token generation. 

Existing LLM solutions are frequently overconfident in their assessments of a text's factual consistency. \citet{tang2024tofueval} showed that their summary-level binary factual consistency prompts, when used with GPT-4, frequently failed to identify sentences with factual errors, resulting in False Positive Rates of 69\% and 46\% on the MediaSum and MeetingBank subsets of the TofuEval Summary-Level dataset, respectively. This issue of overconfidence is not unique to their approach but rather a general problem that neural networks, including LLMs, tend to be overconfident in their predictions \citep{guo2017calibration, minderer2021revisiting, jiang2021know, xiong2023llms}.\footnote{Binary prompts, by their very nature, force models to reduce nuance and uncertainty into a single decision, contributing to the overconfidence observed in their responses.}

Confidence elicitation, an increasingly popular method, involves prompting the LLM to output its uncertainty along with its prediction \citep{lin2022teaching, xiong2023llms, tian2023just}. Despite its potential, no confidence elicitation approach has yet consistently yielded accurate confidence estimates across diverse LLMs and tasks, limiting its current practical utility. Calibration, the process of adjusting model-predicted probabilities to match their empirical accuracies, has remained the leading solution to correct model overconfidence \citep{guo2017calibration}.

We propose a novel approach - crafting a diverse set of LLM prompts that each output a binary score, indicating each prompt's belief of whether the summaries contain any factual errors. These binary features are then fed into ensembling models, which integrate the multiple perspectives of each prompt to produce a single probability. Finally, we calibrate the ensemble models to obtain empirically accurate probabilities that a given summary is factually consistent or free of hallucination.\footnote{We craft prompts that aim to detect either factual inconsistencies or hallucinations, allowing the ensembling model to determine the importance of each prompt for the datasets it is trained on. This approach enables us to sidestep the subtle distinction between factual inconsistencies and hallucinations, and instead let the ensembling model decide what prompts are most relevant for the training data.} The full pipeline of our framework can be seen in Figure \ref{fig:endToEndPipeline}. 

The primary contributions of our work can be outlined as follows: 
\textbf{1:} We demonstrate that prior methods for detecting factual errors in summarizations perform significantly worse without the common practice of an optimized threshold on subsets of the dataset under test.
\textbf{2:} We introduce a large and diverse set of prompts, each employing unique methods and evaluation protocols to detect hallucinations and factual inconsistencies in generated summaries.
\textbf{3:} Our end-to-end framework achieves state-of-the-art balanced accuracy on the AggreFact-XSUM FTSOTA, TofuEval Summary-Level, and HaluEval Summarization benchmarks in detecting factual errors in transformer-generated text summaries, all without fine-tuning the language model or relying on impractical thresholding techniques. 


\section{Datasets}

The AggreFact FTSOTA benchmark \citep{tang2023understanding}  tests a model's ability to identify factual inconsistencies in summaries produced by fine-tuned transformer-based summarization models. The dataset combines nine existing annotated factuality datasets, converting all factual consistency scores to binary. AggreFact is categorized based on the development timeline of the underlying summarization models into FTSOTA, EXFORMER, and OLD categories, and is divided into AggreFact CNN/DM and AggreFact-XSUM subsets. We benchmark AggreFact-XSUM FTSOTA and exclude its CNN/DM counterpart due to its insufficient number of factual inconsistencies and its more extractive summary style, compared to the more abstractive style of current SOTA LLMs.

HaluEval \citep{li2023halueval}, a comprehensive benchmark for assessing hallucination in LLMs, uses a ChatGPT-based two-step sampling-then-filtering framework to create the dataset. We focus on the summarization subset of HaluEval, which pairs each of the 10,000 sampled document contexts with two summaries: one with hallucination and one without.

TofuEval \cite{tang2024tofueval} is a topic-focused dialogue summarization benchmark containing 1.5K LLM-generated summaries from the MediaSum \cite{zhu2021mediasum} and MeetingBank \cite{hu2023meetingbank} datasets. TofuEval uses a two-stage annotation process, where two expert linguists independently assess the binary relevance, completeness, and factual consistency of each sentence in the summaries. We focus solely on the factual consistency annotations of main topic summaries, merging summarization sentences into one paragraph deemed consistent if all summarization sentences are individually marked as consistent.\footnote{\citet{tang2024tofueval} separated summaries into Main and Marginal, with the majority of sentences categorized as Main. We focus on the Main summaries because the Marginal dataset in TofuEval is insufficiently small for reliable analysis.} We thus refer to this concatenation interchangeably as "TofuEval Summary-Level" or simply "TofuEval Summary".

We use the provided test subsets of each dataset, except for HaluEval Summarization, which lacks a train-test split. For HaluEval Summarization, we use a balanced random sample of 3,000 summaries as our test set.

\section{Methodology}

\subsection{Ensembling Methods}

We train and evaluate 16 ensembling methods. When applicable, parameters for all relevant methods are determined using a grid search over the parameter feature space. Below is a listing of these methods:

\begin{itemize}
    \item \textbf{Linear Models}: LogisticRegression, LDA
    \item \textbf{Tree-Based Methods}: RandomForest, GradientBoosting, AdaBoost, DecisionTree, CatBoost, XGB, LGBM
    \item \textbf{Ensemble Voting:} MajorityLabelVoter, WeightedMajorityLabelVoter
    \item \textbf{Label Aggregation Models}: LabelModel, Dawid-Skene\footnote{The LabelModel, as delineated in \citet{Ratner_2017}, is particularly effective in learning the conditional probabilities of labeling functions, adeptly reweighting their outputs in semi-supervised contexts. The LabelModel represents an evolution in semi-supervised learning, encompassing techniques such as those in FlyingSquid \citep{fu2020fast}, Dawid-Skene \citep{10.2307/2346806}, Data Programming \citep{ratner2016data}, and MeTaL \citep{ratner2019training}, all of which were originally included as part of  the Wrench benchmark \citep{zhang2021wrench}.}
    \item \textbf{Other Methods}: Support Vector Machines, Nearest Neighbors, Naive Bayes (BernboulliNB)

\end{itemize}

We consider MajorityLabelVoter as an ensembling baseline, conceptually similar to averaging binary scores with a threshold of 0.5. By comparing the results of the majority vote to other ensembling methods, we can empirically assess the additional performance gains achievable through more sophisticated ensembling techniques.

\subsection{Metrics Used}
Following \citet{tang2023understanding} and \citet{laban2021summac}, we use balanced accuracy to assess each model's proficiency in detecting factual errors. Balanced accuracy helps provide a less biased evaluation by balancing the importance of sensitivity and specificity, ensuring that the predominance of the majority class does not skew the results. 

We measure the reliability of predicted probabilities for factual consistency using Expected Calibration Error (ECE). Expected Calibration Error (ECE) is calculated by partitioning predictions into $n$ bins based on their confidence levels, computing the absolute difference between the actual accuracy and the predicted probability in each bin, and then taking the weighted average of these differences across all bins. We use a ECE effectively measures the discrepancy between a model's confidence in its predictions and its actual performance, with lower values indicating better calibration.  



\begin{equation}
  \text{ECE} = \sum_{m=1}^{M} \frac{|B_m|}{N} \left| \text{acc}(B_m) - \text{conf}(B_m) \right|
\end{equation}

Where: $M$ is the number of bins, $N$ is the total number of samples, $B_m$ is the set of samples in bin $m$, $|B_m|$ is the number of samples in bin $m$, $\text{acc}(B_m)$ is the accuracy of predictions in bin $m$, and $\text{conf}(B_m)$ is the average predicted probability in bin $m$. We selected $M=8$ for all ECE evaluations, as it offers reasonable balance between statistical reliability and resolution.




\subsection{Calibrating Ensembled Models for Reliable Probability Estimates}
Histogram Binning \citep{zadrozny2002transforming}, Bayesian Binning into Quantiles (BBQ) \citep{naeini2015obtaining}, and Isotonic Regression \citep{zadrozny2002transforming} are non-parametric methods methods of calibration, while Temperature Scaling \citep{guo2017calibration} and Platt Scaling \citep{platt1999probabilistic} are parametric. Platt Scaling applies a sigmoid function to model outputs to calibrate them. We test and apply Platt Scaling, BBQ, Histogram Binning, and Isotonic Regression, to obtain reliable probability estimates from our ensembled models.


\subsection{Methodology for Prompt Creation}
Our LLM prompts, created using GPT-4 in the OpenAI playground, employs various Chain of Thought (CoT) \citep{wei2023chainofthought} approaches to guide models through a structured evaluation of factual consistency. Most prompts have various explicit evaluation criteria, requiring the LLM to determine if each claim in the summary can be inferred directly from the context.

\subsection{Selection of Prompt Pool}
Prompt selection for each prompt pool size was determined using Recursive Feature Elimination (RFE) and Minimum Redundancy Maximum Relevance (MRMR) methods \citep{guyon2002gene, ding2005minimum}. Recursive Feature Elimination (RFE) is a feature selection technique that iteratively fits a model and removes the least significant feature based on the model's coefficients or feature importances, until the desired number of features is reached. The Minimum Redundancy Maximum Relevance (mRMR) algorithm iteratively selects features by maximizing their relevance to the target, typically measured through methods like mutual information, while minimizing redundancy among them, often using Pearson correlation.\footnote{We use the \href{https://github.com/smazzanti/mrmr}{mRMR library} and \href{https://scikit-learn.org/stable/modules/generated/sklearn.feature_selection.RFE.html}{Scikit-Learn's RFE} for their respective feature selection methods.} We select the best-performing subset of prompts of these two methods for each prompt size.

\section{Results}

\subsection{Individual Prompt Results}
Table \ref{tab:individual_prompt_performance_across_datasets} displays the top five prompts' performance, highlighting an average 2.5\% increase in balanced accuracy when using GPT-4-Turbo over GPT-3.5-Turbo.\footnote{Preliminary API calling over a smaller sample revealed roughly comparable results with Claude 3 Opus and worse performance with open source LLMs. The use and analysis of other LLMs and open-source models are deferred to future work.}



\begin{table*}[h!]
\centering
\resizebox{\textwidth}{!}{
\begin{tabular}{|l|c|ccc|ccc|ccc|ccc|}
\hline
\multirow{3}{*}{\textbf{Prompt}}& \multicolumn{1}{|c|}{\multirow{3}{*}{\textbf{LLM}}} & \multicolumn{3}{c|}{\textbf{AggreFact-}} & \multicolumn{3}{c|}{\textbf{HaluEval}} & \multicolumn{3}{c|}{\textbf{TofuEval MediaSum}} & \multicolumn{3}{c|}{\textbf{TofuEval MeetingBank}} \\
\multicolumn{1}{|c|}{\multirow{3}{*}{\textbf{}}}& & \multicolumn{3}{c|}{\textbf{XSUM FTSOTA}} & \multicolumn{3}{c|}{\textbf{Summarization}} & \multicolumn{3}{c|}{\textbf{Summary-Level}} & \multicolumn{3}{c|}{\textbf{Summary-Level}} \\
\cline{3-14}
& & \textbf{Bal. Acc.} & \textbf{Precision} & \textbf{Recall} & \textbf{Bal. Acc.} & \textbf{Precision} & \textbf{Recall} & \textbf{Bal. Acc.} & \textbf{Precision} & \textbf{Recall} & \textbf{Bal. Acc.} & \textbf{Precision} & \textbf{Recall} \\
\hline
\multirow{2}{*}{Prompt 1} & GPT-3.5 & 67.4 & 64.5 & 81.8 & 69.7 & 62.8 & 96.3 & 57.6 & 61.4 & 84.1 & 67.0 & 73.9 & 87.0 \\
                          & GPT-4   & \textbf{69.1} & 65.9 & \textbf{82.8} & 71.5 & 64.3 & \textbf{96.5} & 60.2 & 63.2 & 84.1 & \textbf{73.2} & 78.4 & 88.2 \\
\hline
\multirow{2}{*}{Prompt 2} & GPT-3.5 & 66.4 & 63.8 & 80.4 & 67.2 & 62.8 & 96.3 & 58.4 & 61.2 & 96.0 & 61.0 & 69.3 & 93.5 \\
                          & GPT-4   & 68.4 & 65.5 & 81.4 & 68.7 & 64.2 & 84.6 & 61.4 & 63.0 & \textbf{96.0} & 70.3 & 75.1 & \textbf{94.7} \\
\hline
\multirow{2}{*}{Prompt 3} & GPT-3.5 & 65.0 & 71.6 & 51.2 & 70.2 & 67.0 & 79.3 & 63.2 & 65.5 & 84.1 & 60.0 & 70.2 & 75.1 \\
                          & GPT-4   & 66.8 & \textbf{73.8} & 53.3 & \textbf{72.7} & \textbf{69.3} & 81.3 & 63.9 & 66.0 & 84.8 & 64.3 & 73.1 & 78.7 \\
\hline
\multirow{2}{*}{Prompt 4} & GPT-3.5 & 61.7 & 60.0 & 76.8 & 63.7 & 61.4 & 73.9 & 65.4 & 70.0 & 69.5 & 68.7 & 77.4 & 75.1 \\
                          & GPT-4   & 62.8 & 60.9 & 77.5 & 66.4 & 63.7 & 75.9 & \textbf{66.6} & \textbf{71.1} & 70.2 & 72.0 & \textbf{79.6} & 78.7 \\
\hline
\multirow{2}{*}{Prompt 5} & GPT-3.5 & 60.2 & 58.8 & 76.5 & 66.4 & 62.8 & 80.4 & 61.0 & 63.8 & 84.1 & 63.3 & 71.9 & 81.7 \\
                          & GPT-4   & 61.8 & 60.1 & 77.5 & 68.2 & 64.4 & 81.6 & 61.8 & 64.3 & 84.8 & 67.6 & 74.2 & 83.2 \\
\hline
\end{tabular}
}
\caption{Individual performance of the top-five performing prompts across all test datasets. GPT-4 refers to GPT-4-Turbo (gpt-4-0125-preview), while GPT-3.5 refers to GPT-3.5-Turbo (gpt-3.5-turbo-1106) for all datasets except HaluEval Summarization, where GPT-3.5-Turbo-16K (gpt-3.5-turbo-16k-0613) is used due to the longer context length. The best performing individual prompt-model-metric combination for each dataset is shown in bold. Prompts generally exhibit higher recall than precision, indicating a higher risk of failing to identify factual inconsistencies or hallucinations in generated text compared to misclassifying an accurate summary as erroneous.
}
  \label{tab:individual_prompt_performance_across_datasets}
\end{table*}

\subsection{Ensemble Benchmarking and Comparing to Existing Methods}
    

For each test dataset, we train ensemble models on the binary LLM prompt outputs using only the three remaining non-test datasets. Table \ref{tab:ensemble_benchmarking} shows the impact of different LLM prompt sizes and ensembling methods on balanced accuracy for the AggreFact-XSUM FTSOTA, HaluEval Summarization, and TofuEval Summary-Level test datasets.\footnote{Table \ref{tab:ensemble_benchmarking} reports pre-calibration performance for ensembling results. Calibration had minimal effect on balanced accuracy, ranging from 0\% to 0.2\%.} The LabelModel \citep{Ratner_2017} is the most frequent top performer among the ensemble models across dataset and prompt size combinations. Snorkel's LabelModel emerges as the top performer among the ensemble models, likely because it is designed to effectively combine weak and noisy classifiers, which accurately describe the individual performance of the LLM prompts. 

Table \ref{tab:ensemble_benchmarking} demonstrates that ensembling binary outputs from multiple prompts can substantially improve performance compared to the best individual LLM prompts, shown in Table \ref{tab:individual_prompt_performance_across_datasets}. Increasing the number of ensembled prompts from 5 to 9 does not consistently improve performance, likely because the top-5 prompts use GPT-4 while the remaining prompts use GPT-3.5. The limited training data may hinder the ensembling models' ability to benefit from the less reliable GPT-3.5 prompts.

\begin{table*}[h!]
\centering
\resizebox{\textwidth}{!}{
\begin{tabular}{|l|ccc|ccc|ccc|ccc|}
\hline
\multicolumn{1}{|c|}{\multirow{3}{*}{\textbf{Ensembling}}} & \multicolumn{3}{c|}{\textbf{AggreFact-}} & \multicolumn{3}{c|}{\textbf{HaluEval}} & \multicolumn{3}{c|}{\textbf{TofuEval MediaSum}} & \multicolumn{3}{c|}{\textbf{TofuEval MeetingBank}} \\
\multicolumn{1}{|c|}{\multirow{3}{*}{\textbf{Method}}}& \multicolumn{3}{c|}{\textbf{XSUM FTSOTA}} & \multicolumn{3}{c|}{\textbf{Summarization}} & \multicolumn{3}{c|}{\textbf{Summary-Level}} & \multicolumn{3}{c|}{\textbf{Summary-Level}} \\
\cline{2-13}
\multicolumn{1}{|c|}{} & \multicolumn{3}{c|}{\textbf{Number of Prompts}} & \multicolumn{3}{c|}{\textbf{Number of Prompts}} & \multicolumn{3}{c|}{\textbf{Number of Prompts}} & \multicolumn{3}{c|}{\textbf{Number of Prompts}} \\
& \textbf{3} & \textbf{5} & \textbf{9} & \textbf{3} & \textbf{5} & \textbf{9} & \textbf{3} & \textbf{5} & \textbf{9} & \textbf{3} & \textbf{5} & \textbf{9} \\
\hline

Baseline & 50.00 & 50.00 & 50.00 & 50.00 & 50.00 & 50.00 & 50.00 & 50.00 & 50.00 & 50.00 & 50.00 & 50.00 \\
AdaBoost & 67.04 & 67.04 & 70.21 & 71.47 & 73.73 & 72.67 & \textbf{65.80} & 63.80 & 64.74 & 71.17 & 72.89 & 71.97 \\
BernoulliNB & 67.04 & 71.02 & 68.07 & 73.27 & 72.33 & 68.57 & 64.24 & 65.96 & 64.80 & 74.07 & 76.38 & 77.16 \\
CatBoost & 67.04 & 67.04 & 68.73& 71.47 & 71.80 & 71.53 & \textbf{65.80} & 63.80 & 64.74 & 71.17 & 72.89 & 73.80 \\
DawidSkene & 69.17 & 69.62 & 70.28 & 73.27 & 70.53 & 67.37 & 64.17 & 64.33 & 63.47 & 74.07 & 73.70 & 71.91 \\
DecisionTree & 67.04 & 67.04 & 66.11 & 73.27 & 68.67 & 71.50 & 60.16 & 64.90 & 63.31 & 73.16 & 73.91 & 67.00 \\
GradientBoosting & 67.04 & 66.53 & 68.91 & 73.90 & 71.07 & 67.63 & \textbf{65.80} & 64.67 & 64.30 & 71.17 & 73.91 & 73.80 \\
KNeighbors & 67.04 & 67.04 & 67.39 & \textbf{74.87} & 71.43 & 68.77 & 64.24 & 64.17 & 63.28 & 74.07 & 69.55 & 71.08 \\
LDA & 67.04 & 67.04 & 69.79 & 73.27 & 73.83 & 70.60 & \textbf{65.80} & 63.80 & 64.44 & 71.17 & 72.89 & 71.03 \\
LabelModel &\textbf{69.40} & \textbf{71.92} & 67.93 & 73.90 & 71.67 & 68.50 & 64.24 & \textbf{66.33} & \textbf{65.53} & \textbf{74.10} & \textbf{79.38} & \textbf{79.74} \\
LGBM & 67.04 & 67.04 & 68.73 & 73.90 & \textbf{74.07} & \textbf{73.43} & \textbf{65.80} & 64.67 & 65.17 & 74.07 & 73.91 & 73.58 \\
LogisticRegression & 67.04 & 67.04 & 70.17 & 73.27 & 72.53 & 70.40 & \textbf{65.80} & 63.80 & 63.77 & 71.17 & 72.89 & 70.74 \\
MajorityLabelVoter & 69.17 & 69.62 & \textbf{71.05} & 73.27 & 70.37 & 67.87& 64.17 & 64.33 & 63.50 & \textbf{74.10}& 73.70 & 70.67 \\
RandomForest & 67.04 & 66.72 & 67.91 & 73.90 & 70.87 & 70.77 & \textbf{65.80} & 64.90 & 63.54 & 71.17 & 73.99 & 73.58 \\
SVC & 67.04 & 66.72 & 67.61 & 73.90 & 72.87 & 72.03 & \textbf{65.80} & 64.67 & 64.30 & 71.17 & 73.91 & 72.27 \\
WeightedMajorityVoting & 67.04 & 67.04 & 70.17 & 73.27 & 73.73 & 72.60 & \textbf{65.80} & 63.80 & 63.77 & 71.17 & 72.89 & 70.74 \\
XGB & 67.04 & 66.72 & 70.39 &73.90 & 71.73 & 66.73 & \textbf{65.80} & 64.90 & 63.87 & 71.17 & 73.99 & 72.14 \\
\hline

\end{tabular}
}
\caption{Exploring the impact of various LLM prompt sizes and ensembling methods on balanced accuracy across all test datasets. The top performing ensemble method for each prompt size-dataset combination is shown in bold. Ensembling as few as three prompts consistently yields performance improvements across all datasets compared to the best-performing individual prompt.}

\label{tab:ensemble_benchmarking}

\end{table*}

Table \ref{tab:sota_comparison} compares the balanced accuracy of state-of-the-art encoder-based factual consistency models and LLM solutions to our ensemble methods in identifying factual inconsistencies and hallucinations. As shown in Table \ref{tab:sota_comparison}, our proposed ensemble approach outperforms all existing methods across the benchmark datasets.\footnote{The QAFactEval and QuestEval results for HaluEval Summarization were obtained using a balanced random sample of 2,000 context-summary pairs. LLM Solution method results on HaluEval were obtained using a slightly larger balanced random sample of 3,000 context-summary pairs. All other results use the full datasets.}

\begin{table*}[ht!]
  \centering
  \resizebox{\textwidth}{!}{
    \begin{tabular}{|c|l|c|c|c|c|}
    \hline
    \multicolumn{1}{|c|}{\multirow{2}{*}{\textbf{Method Type}}} & \multicolumn{1}{|c|}{\multirow{2}{*}{\textbf{Method}}} & \multicolumn{1}{|c|}{\multirow{2}{*}{\textbf{AggreFact-}}} & \multicolumn{1}{|c|}{\multirow{2}{*}{\textbf{HaluEval}}} & \multicolumn{1}{c|}{\multirow{2}{*}{\textbf{TofuEval MediaSum}}} & \multicolumn{1}{c|}{\multirow{2}{*}{\textbf{TofuEval MeetingBank}}} \\
    \centering & \centering & & & & \\[-6pt]
    \centering & \centering & \textbf{XSUM FTSOTA} & \textbf{Summarization} & \textbf{Summary-Level} & \textbf{Summary-Level} \\
    \hline
        \multirow{4}{*}{Encoder Models} &AlignScore & 70.2 $\pm$ 3.8 & 65.8 $\pm$ 0.7 & 62.3 $\pm$ 5.8 & 70.1 $\pm$ 5.4\\
        &QuestEval  & 51.3 $\pm$ 4.2 & 55.4 $\pm$ 2.2 & 53.7 $\pm$ 6.0 & 50.8 $\pm$ 5.9\\
        &SummaC-ZS & 53.0 $\pm$ 4.2 & 60.5 $\pm$ 0.7 & 48.6 $\pm$ 6.0 & 57.4 $\pm$ 5.9 \\
        &SummaC-Cv& 50.9 $\pm$ 4.2 & 58.7 $\pm$ 0.7& 53.3 $\pm$ 5.9 & 45.2 $\pm$ 6.1 \\
        &QAFactEval & 62.4 $\pm$ 4.0 &52.0 $\pm$ 2.2 & 54.6 $\pm$ 5.9 & 57.8 $\pm$ 5.7\\
        \hline
        \multirow{6}{*}{LLM Solutions} &ChatGPT-ZS (GPT-3.5)& 62.1 $\pm$ 4.0 & 64.3 $\pm$ 1.8 & 63.6 $\pm$ 5.8 & 69.0 $\pm$ 5.3\\
        &ChatGPT-COT (GPT-3.5)& 55.4 $\pm$ 4.1 & 62.5 $\pm$ 1.8 & 63.1 $\pm$ 5.8 & 66.8 $\pm$ 5.2\\
        &ChatGPT-DA (GPT-3.5)& 56.4 $\pm$ 4.1 & 59.6 $\pm$ 1.8 & 52.7 $\pm$  5.9 & 52.3 $\pm$ 5.8\\
        &ChatGPT-Star (GPT-3.5)& 55.4 $\pm$ 4.1 & 61.5 $\pm$ 1.8 & 57.4 $\pm$ 5.8 & 55.7 $\pm$ 5.7\\
        &Tang2024-Summary (GPT-3.5) & 62.4  $\pm$ 4.0 & 64.0 $\pm$ 0.9 & 61.9 $\pm$ 3.4  & 71.9 $\pm$ 3.1 \\
        &Tang2024-Summary (GPT-4) & 62.9 $\pm$ 4.0 & 66.1 $\pm$ 0.9 & 62.3 $\pm$ 3.4 &  72.9 $\pm$ 3.1\\
        \hline
        \multirow{3}{*}{LLM Ensembles}
        &Ensemble-Top-3 (GPT-4) & 69.4 $\pm$ 3.8 &\textbf{74.9} $\pm$ 1.6  & 65.8 $\pm$ 5.7 & 74.1 $\pm$ 5.1\\
        &Ensembled-Top-5 (GPT-4)&\textbf{71.9}$ \pm$ 3.8 & 74.1 $\pm$ 1.6 & \textbf{66.3}  $\pm$ 5.7& 79.4 $\pm$ 5.1\\
        &Ensemble-Top-9 (Mixed)& 71.1 $\pm$ 3.8 & 73.4 $\pm$ 1.6  &65.5 $\pm$ 5.7 & \textbf{79.7} $\pm$ 4.9\\
            \bottomrule
        \end{tabular}
        }
    \caption{A chart comparing the balanced accuracy of encoder-based models and LLM solutions in identifying factual inconsistencies and hallucinations. For each test dataset, the encoder-based model scores are obtained using linear thresholds optimized on the other three datasets, ensuring that neither the test data nor its validation set is used for threshold tuning. Lower performance in encoder models compared to existing studies is due to evaluating without fine-tuning each model's threshold using the test dataset's development subset. 95\% confidence intervals are shown, with the highest performing method for each dataset in bold. Mixed refers to ensembling the binary LLM outputs from both GPT-3.5 and GPT-4. The existing prompts and relevant citations use in Method Type LLM Solutions can be found in Appendix Section \ref{existing_prompts}.}
    \label{tab:sota_comparison}
\end{table*}

\begin{table*}[h!]
  \centering
  \resizebox{\textwidth}{!}{
    \begin{tabular}{|l|cc|cc|cc|}
    \hline
    \multicolumn{1}{|c|}{\multirow{3}{*}{\textbf{Model}}}
 & \multicolumn{6}{|c|}{\textbf{AggreFact-XSUM FTSOTA}}  \\
    \cline{2-7}
    \multicolumn{1}{|c|}{} &  \multicolumn{2}{|c|}{\textbf{3 Prompts}} & \multicolumn{2}{|c|}{\textbf{5 Prompts}} & \multicolumn{2}{|c|}{\textbf{9 Prompts}} \\ 
    \cline{2-7}
     & Uncal. (\%) & Platt (\%) & Uncal. (\%) & Platt (\%) & Uncal. (\%) & Platt (\%) \\ 
\hline
AdaBoost & 5.1 & 2.1 & 12.0 & 5.7 & 18.7 & 4.3  \\
BernoulliNB & 9.2 & 5.1 & 15.9 & 5.0 & 21.5 & 7.2 \\
CatBoost & 8.2 & 2.5 & 8.1 & 5.2 & 7.7 & 6.7  \\
DecisionTree & 8.9 & 6.4 & 7.2 & 5.3 & 9.3 & 4.9 \\
GradientBoosting & 7.5 & 4.6 & 6.1 & 5.2 & 8.6 & 5.4  \\
KNeighbors & 11.0 & 4.9 & 10.4 & 4.7 & 9.8 & 4.5  \\
LabelModel & 14.7 & 4.1 & 23.8 & 4.7 & 22.8 & 5.3  \\
LDA & 6.9 & 4.5 & 7.4 & 6.1 & 7.1 & 4.1 \\
LGBM & 15.5 & 4.9 & 20.0 & 7.7 & 15.9 & 6.1 \\
LogisticRegression & 13.4 & 4.9 & 6.5 & 5.8 & 6.6 & 3.9  \\
MultinomialNB & 8.3 & 8.1 & 4.2 & \textbf{2.6} & 4.4 & \textbf{2.6} \\
RandomForest & 9.8 & 6.3 & 6.0 & 4.1 & 9.8 & 4.5  \\
SVC & 9.3 & \textbf{0.9} & 8.4 & 6.5 & 9.5 & 5.7 \\
XGB & 7.2 & 6.4 & 6.9 & 4.8 & 16.0 & 6.9  \\
    \bottomrule
    \end{tabular}%
    }
    \caption{Comparison of the Expected Calibration Error (ECE) for ensembling models before (uncalibrated) and after calibration using Platt Scaling, across various models and prompt pools. The calibration model was trained only on the three non-test datasets.}
    \label{tab:calibration_results}
\end{table*}

\subsection{Calibration Analysis}
We investigate the effectiveness of applying calibration to obtain reliable probability estimates from our ensembled models. Table~\ref{tab:calibration_results} shows that applying Platt Scaling significantly reduces Expected Calibration Error (ECE) across various ensembling techniques. Appendix Table~\ref{tab:calibration_results_complete} provides ECE scores before and after calibration across all four datasets including other popular calibration methods such as BBQ, Histogram Binning, and Isotonic Regression. As seen in Table~\ref{tab:calibration_results_complete}, Platt Scaling \citep{platt1999probabilistic} generally outperforms the other tested calibration methods, achieving post-calibration ECE scores of under 7\%.

\begin{figure*}[ht!] 
  \centering
  \includegraphics[width=\textwidth]{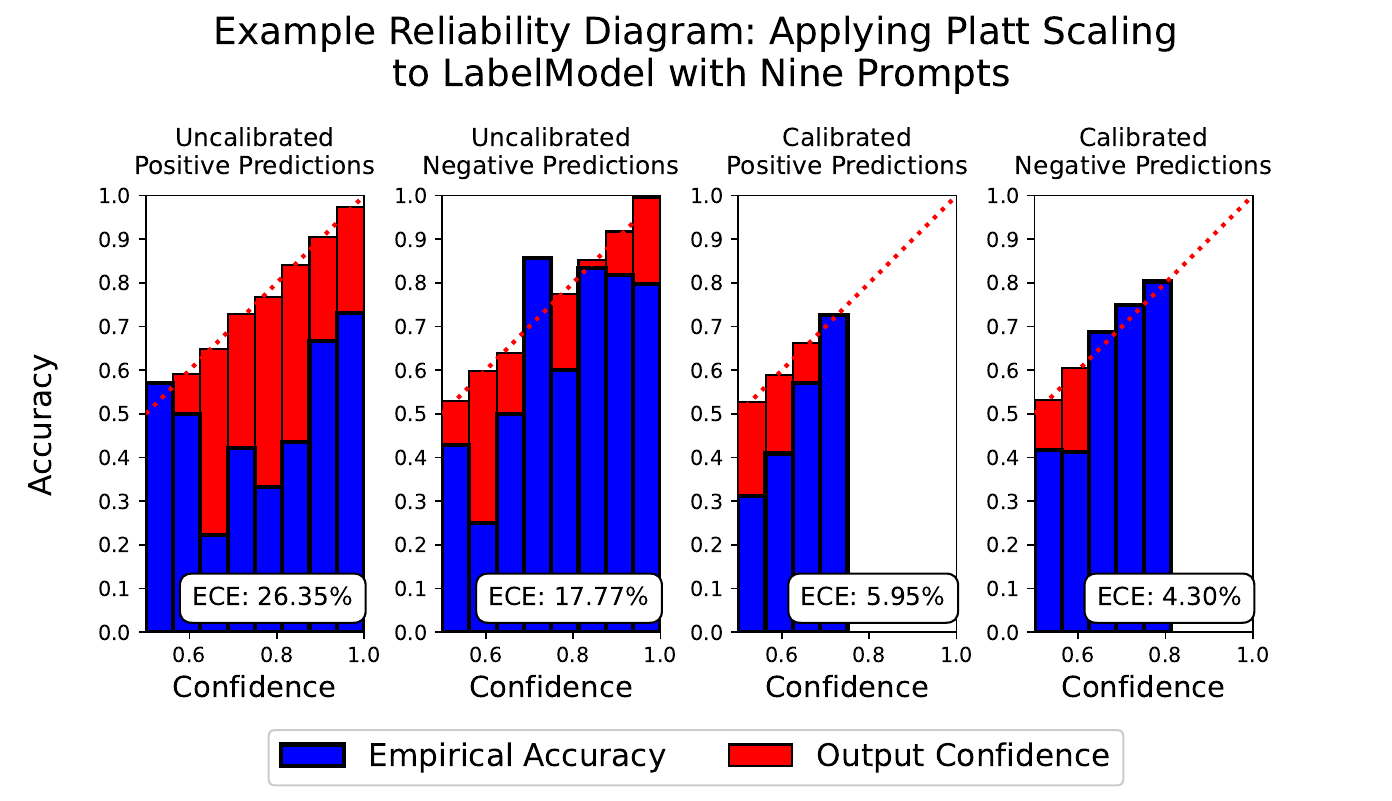} 
  \caption{An example reliability diagram highlighting the difference in reliability between the predicted probabilities before and after calibration. The reliability diagram is separated by positive and negative predictions, highlighting the contrast in model confidences between predicting whether a summary is factually consistent with the source context versus inconsistent. In this visualization, LabelModel was trained with the output of nine LLM Prompts on the non-test datasets and tested on the AggreFact-XSUM FTSOTA test dataset.
  }
  \label{fig:figure_showing_calibrated_model}
\end{figure*}

Reliability diagrams are tools for assessing the accuracy of probability estimates from both uncalibrated and calibrated ensemble models. The reliability diagram in Figure \ref{fig:figure_showing_calibrated_model} shows that Platt Scaling, when applied to an ensemble model, can significantly reduce overconfidence in predicting a text's factual consistency.

\subsection{Statistical Testing}

We assess if the performance gains in LLM ensembling over existing methods in models are statistically significant. For the AggreFact-XSUM FTSOTA, TofuEval, and HaluEval Summarization dataset, we execute distinct evaluations using bootstrap resampling techniques \citep{effron2022}, comparing the best Ensembling Method against the previous top performing model for each dataset. Following \citet{laban2021summac}, we conduct comparisons at a statistical significance level of $p = 0.01$, incorporating the Bonferroni adjustment \citep{Bonferroni1935} due to the multiple tests conducted on the datasets. Our LLM ensembling methods demonstrate statistically meaningful advancements on the HaluEval dataset, with statistically significant p-values of under .01 in both the original and Bonferroni-adjusted analyses. However, the statistical tests do not confirm a statistically significant advantage, with p-values below .01, compared to the second-best methods across the remaining three datasets. The limited number of samples in the AggreFact-XSUM FTSOTA and TofuEval datasets, in contrast to the larger HaluEval dataset, necessitates a significantly greater improvement in performance to achieve statistical significance. \footnote{To eliminate any biases caused by varying sample sizes across models, we exclusively conduct statistical testing on the overlapping subsets of all samples under consideration, thereby ensuring that differing sample sizes do not affect our statistical significance testing.}

\section{Conclusion}



We introduce \textit{Detecting Errors through Ensembling Prompts (DEEP)}, a state-of-the-art LLM-based method for detecting factual consistencies and hallucinations in summaries. Our findings reveal that factual consistency encoder models exhibit a pronounced sensitivity to threshold settings, with their performance markedly declining if the threshold is not adjusted based on the test datasets. We demonstrate that \textit{DEEP} surpasses the performance of existing methods and models in evaluating the factual consistency of summaries produced by recent transformer models. Finally, we show that by calibrating ensemble models using binary input features derived from LLM prompts, \textit{DEEP} achieves reliable probabilities indicating a text's factual consistency or presence of errors.

\section*{Limitations}

\alexnote{mention fine-tuning LLMs for factual inconsistency and hallucination detection}
LLM approaches for identifying factual errors are significantly more resource-intensive than existing fine-tuned encoder models, requiring up to three orders of magnitude more parameters. Our method requires more computation; however, this cost may be justified in high-stakes situations where failing to identify factual errors is costly.


Future research should explore the performance of these prompts on more powerful language models as they become available. Additionally, future work should create a dataset of quality examples of chain-of-thought reasoning to identify factual errors, enabling few-shot learning to boost LLM performance. Future work should consider comparing the performance of ensembling factual consistency model scores to LLM prompt ensembling.

Future work should investigate why encoder models for factual consistency evaluation require dataset-specific linear thresholding for optimal performance. Possible reasons include: (1) the dependence of optimal binary classification thresholds on the average number of sentences per summary, as most models calculate summary factuality scores by averaging sentence-level scores; (2) dataset imbalances in the ratio of summaries with factual errors, with thresholding balancing specificity and sensitivity; and (3) inherent differences in identifying factual errors across datasets, even when controlling for summary length and annotation imbalances, due to factors like summarization models or the context's source. However, these reasons are speculative, and further experiments are needed to test these hypotheses and explore approaches that reduce model sensitivity to dataset characteristics.





It is uncertain how many labeling errors exist in the AggreFact-XSUM-SOTA and HaluEval Summarization datasets. Future datasets should create summarization error datasets, where, like TofuEval, ground truth labels are generated by combining multiple humans annotations. Additionally, the datasets under test contained solely English Summaries. To evaluate LLMs' ability to identify errors across languages, there is a need for multi-lingual summarization error datasets. 

Fine-tuning LLMs for error detection and applying our end-to-end pipeline for spotting wider ranges of language generation errors, including those in QA and machine translation, should be explored.

\alexnote{commented out section on why encoder models require linear thresholding per evaluation dataset. will probably include this in a follow up work.}
\alexnote{The Acknowledgement section is currently commented out so that we keep our paper double-blind. However, the plan is to reinclude this in the papers publication.}
\bibliography{main}

\appendix
\label{sec:appendix}

\section{Our Prompts}

\begin{tcolorbox}[colback=white,colframe=blue!75!black,title=Prompt 1, breakable]
Assess the factual consistency of the following claim based solely on the information provided in the summary. Do not make inferences or assumptions beyond the provided information.

Summary: \{...\}\\
Claim: \{...\}\\
Assessment:
Think step by step. Use the following chain of thought.\\
Step 1: Remember the claim.\\
Step 2: Remember the summary.\\
Step 3: Break down all the lines in the claim and remember them. Pay close attention to numbers associated with the events and store them separately.\\
Step 4: Follow the given chain of thought.\\
After following these steps, decide if the claim is SUPPORTED or NOT SUPPORTED. If you are not sure, output NOT SUPPORTED. Only output SUPPORTED if you are 100 percent confident that the claim is fully supported by the summary. But regardless, be sure that you output your answer.
\end{tcolorbox}

\begin{tcolorbox}[colback=white,colframe=blue!75!black,title=Prompt 2, breakable]
Assess the factual consistency of the following claim based solely on the information provided in the summary. Do not make inferences or assumptions beyond the provided information.

Summary: \{...\}\\
Claim: \{...\}\\
Assessment:
Think step by step. Use the following chain of thought.\\
Step 1: Remember the claim.\\
Step 2: Remember the summary.\\
Step 3: Break down all the lines in the claim and remember them. Pay close attention to numbers associated with the events and store them separately.\\
Step 4: Use multiple factuality metrics (e.g., ChatGPT-based metrics, FEQA, QAGS) to evaluate the factuality of every line. Compare and contrast the results from each metric, discussing any variations in their effectiveness in detecting factual errors. Explain which metric appears to be most effective and why, based on the specific characteristics of the summary.\\
Step 5: Summarize results.\\
Step 6: Be strict. Even if one line is factually inconsistent, mark as UNSUPPORTED; else mark as SUPPORTED.\\
\end{tcolorbox}

\begin{tcolorbox}[colback=white,colframe=blue!75!black,title=Prompt 3, breakable]
In this task, you are required to analyze the factual consistency of a summary against the original article by directly comparing key points.\\
    Procedure:\\
    1. Identify Key Points:
       - List the key points and claims made in the summary and the corresponding points in the article.\\
    2. Comparative Analysis:
       - Create a side-by-side comparison for each key point between the summary and the article.
       - Note any discrepancies, no matter how minor, in each comparison.\\
    3. Detailed Discrepancy Evaluation:
       - For each noted discrepancy, determine whether it falls under any specific error type (e.g., Negation, Adjective, Coreference, etc.).
       - Evaluate the impact of each discrepancy on the overall factual consistency.
       - For each noted discrepancy, determine whether it falls under any specific error type (e.g., Adjective Error, Coreference Error, Number Error, Entity Error, Attribute Error, Pronoun Error, Commonsense Error, Temporal Error, Predicate Error, Discourse Link Error, Relation Error, Quantity Error, Event Error, Noun Phrase Error, Circumstance Error, Hallucination Error).
       - ENSURE THAT YOU THOROUGHLY AND COMPREHENSIVELY CHECK FOR ANY Adjective Error, Coreference Error, Number Error, Entity Error, Attribute Error, Pronoun Error, Commonsense Error, Temporal Error, Predicate Error, Discourse Link Error, Relation Error, Quantity Error, Event Error, Noun Phrase Error, Circumstance Error, Hallucination Error
       - If any of the following discrepancies fall under the listed error types, output NOT SUPPORTED.\\
    4. Strict Criteria for Comparative Support:
       - Classify the summary as 'SUPPORTED' only if there are no discrepancies in the comparative analysis.
       - If any discrepancy is found, classify the summary as 'NOT SUPPORTED'.\\
    Article: \{...\}\\
    Summary: \{...\}\\
    Answer (SUPPORTED or NOT SUPPORTED):
\end{tcolorbox}

\section{Existing Prompts}
\label{existing_prompts}
\begin{tcolorbox}[colback=white, colframe=pink!, title=\textcolor{black}{\citet{li2023halueval} Prompt For Identifying Factual Consistencies}, breakable]
Follow the instructions. \#\#\# Instruction: Determine if the text is consistent or inconsistent with the provided knowledge and dialogue history. If there is a logical conflict, respond with 'inconsistent'. If there is no conflict, respond with 'consistent'.\\
Input: Text: \{...\}.\\
Dialogue History: \{...\}: Please summarize the given knowledge.\\
Response: 
\end{tcolorbox}

\label{Luo 2023 Zero-Shot Prompt}
\begin{tcolorbox}[colback=white,colframe=pink!,title={\textcolor{black}{\citet{luo2023chatgpt} Zero-Shot}}, breakable]
Decide if the following summary is consistent with the corresponding article. Note that consistency means all information in the summary is supported by the article.\\
Article: [...]\\
Summary: [...]\\
Answer (yes or no):
\end{tcolorbox}


\begin{tcolorbox}[colback=white,colframe=pink!,title={\textcolor{black}{\citet{luo2023chatgpt} Chain-of-Thought}}, breakable]
Decide if the following summary is consistent with the corresponding article. Note that consistency means all information in the summary is supported by the article.\\
Article: [...]\\
Summary: [...]\\
Explain your reasoning step by step then answer (yes or no) the question:
\end{tcolorbox}

\begin{tcolorbox}[colback=white,colframe=pink!,title={\textcolor{black}{\citet{wang2023chatgpt} ChatGPT-DA}}, breakable]
Score the following news summarization given the corresponding news with respect to consistency on a continuous scale from 0 to 100, where a score of zero means “inconsistency” and score of one hundred means “perfect consistency”. Note that consistency measures whether the facts in the summary are consistent with the facts in the original article. Consider whether the summary does reproduce all facts accurately and does not make up untrue information.\\
Article: [...]\\
Summary: [...]\\
Scores:
\end{tcolorbox}

\begin{tcolorbox}[colback=white,colframe=pink,title={\textcolor{black}{\citet{wang2023chatgpt} ChatGPT-Star}}, breakable]
Score the following news summarization given the corresponding news with respect to consistency with one to five stars, where one star means “inconsistency” and five stars means “perfect consistency”. Note that consistency measures whether the facts in the summary are consistent with the facts in the original article. Consider whether the summary does reproduce all facts accurately and does not make up untrue information.\\
Article: [...]\\
Summary: [...]\\
Stars:
\end{tcolorbox}

\section{Supplemental Results}
Table~\ref{tab:calibration_results_complete} provide the full calibration results before and after calibration across all datasets. Table~\ref{tab:comprehensive_performance_summary} shows the performance gained through ensembling compared to using the best performing individual prompt.

\begin{table*}[!htb]
\centering
\resizebox{\textwidth}{!}{
\begin{tabular}{|l|ccccc|ccccc|ccccc|ccccc|}
\hline
\multicolumn{1}{|c|}{\multirow{2}{*}{\textbf{Model}}} & \multicolumn{5}{|c|}{\textbf{AggreFact-XSUM FTSOTA}} & \multicolumn{5}{|c|}{\textbf{AggreFact-CNN/DM FTSOTA}} & \multicolumn{5}{|c|}{\textbf{TofuEval MediaSum}} & \multicolumn{5}{|c|}{\textbf{TofuEval MeetingBank}} \\
\cline{2-21}
& Uncal. & Platt & BBQ & Hist. & Isotonic & Uncal. & Platt & BBQ & Hist. & Isotonic & Uncal. & Platt & BBQ & Hist. & Isotonic & Uncal. & Platt & BBQ & Hist. & Isotonic \\
\hline
AdaBoost & 5.1 & 2.1 & 3.5 & 3.7 & 3.7 & 11.2 & 6.9 & 8.6 & 7.4 & 5.7 & 12.6 & 8.2 & 6.3 & 11.0 & 10.9 & 5.7 & 6.7 & 7.7 & 9.5 & 8.7 \\
BernoulliNB & 9.2 & 5.1 & 4.8 & 4.8 & 4.2 & 11.8 & 7.2 & 10.6 & 8.1 & 6.1 & 7.3 & 10.3 & 7.1 & 7.8 & 8.3 & 14.8 & 4.6 & 1.9 & \textbf{1.7} & 2.9 \\
CatBoost & 8.2 & 2.5 & 4.2 & 5.8 & 5.2 & 6.1 & 7.4 & 9.1 & 9.1 & 9.1 & 9.8 & 12.8 & 12.8 & 7.3 & 8.5 & 6.8 & 6.7 & 5.3 & 9.8 & 5.4 \\
DecisionTree & 8.9 & 6.4 & 6.6 & 6.6 & 4.6 & 3.9 & \textbf{2.9} & 4.9 & 7.5 & 3.9 & 5.8 & 10.8 & 8.0 & 8.9 & 5.8 & 5.2 & 9.9 & 8.2 & 7.7 & 5.2 \\
GradientBoosting & 7.5 & 4.6 & 5.8 & 7.1 & 4.1 & 6.3 & 9.3 & 9.8 & 9.3 & 7.3 & 9.8 & 11.2 & 8.9 & 9.1 & 8.7 & 5.3 & 9.4 & 2.6 & 9.8 & 6.3 \\
KNeighbors & 11.0 & 4.9 & 13.2 & 7.5 & 4.5 & 5.4 & 8.4 & 7.6 & 6.3 & 7.4 & 6.9 & 9.4 & 14.5 & 9.0 & 7.8 & 7.0 & 5.4 & 8.0 & 5.7 & 4.8 \\
LabelModel & 14.7 & 4.1 & 4.3 & 8.4 & 6.2 & 15.9 & 8.5 & 9.9 & 5.6 & 7.3 & 11.8 & 14.9 & 16.3 & 9.7 & 13.7 & 18.8 & 1.9 & 6.2 & 2.9 & 5.4 \\
LDA & 6.9 & 4.5 & 4.5 & 3.6 & 3.4 & 6.8 & 6.1 & 9.8 & 7.8 & 6.4 & 8.5 & 11.8 & 6.3 & 11.0 & 10.8 & 7.6 & 5.5 & 4.5 & 8.0 & 4.3 \\
LGBM & 15.5 & 4.9 & 6.0 & \textbf{5.8} & 5.3 & 11.0 & 6.0 & 7.3 & 6.2 & 5.5 & 13.1 & 10.3 & 11.3 & 10.7 & 9.5 & 7.3 & 4.9 & 3.2 & 6.5 & 5.6 \\
LogisticRegression & 13.4 & 4.9 & 5.7 & 5.7 & 5.7 & 5.7 & 6.7 & 9.6 & 8.5 & 6.3 & 10.8 & 11.1 & 6.1 & 10.9 & 9.8 & 4.3 & 5.4 & 7.3 & 7.3 & 7.3 \\
MultinomialNB & 8.3 & 8.1 & 7.8 & 7.9 & 7.9 & 12.9 & 6.8 & 10.2 & 9.0 & 6.7 & 10.1 & \textbf{4.7} & 5.0 & 7.1 & 6.7 & 17.7 & 5.4 & 4.9 & 3.2 & 4.0 \\
RandomForest & 9.8 & 6.3 & 7.0 & 9.9 & 6.0 & 10.9 & 12.6 & 15.4 & 10.5 & 10.0 & 13.2 & 19.7 & 17.3 & 12.8 & 13.0 & 8.4 & 16.3 & 6.3 & 14.7 & 9.2 \\
SVC & 9.3 & \textbf{0.9} & 11.1 & 5.7 & 8.3 & 5.8 & 10.7 & 16.9 & 11.1 & 7.8 & 13.4 & 14.8 & 11.7 & 13.5 & 11.0 & 5.4 & 10.5 & 11.1 & 8.1 & 5.3 \\
XGB & 7.2 & 6.4 & 8.8 & 12.4 & 6.5 & 11.9 & 6.9 & 23.0 & 16.7 & 12.4 & 14.5 & 8.2 & 14.4 & 22.4 & 12.9 & 8.2 & 4.9 & 6.9 & 8.1 & 6.9 \\
\hline
\end{tabular}%
}
\caption{ECE comparison for ensembled models: uncalibrated vs. calibrated with Platt Scaling, BBQ, Histogram Binning, and Isotonic across datasets. Highlighted text denotes the ensemble-calibration pair with the lowest ECE per dataset.}
\label{tab:calibration_results_complete}
\end{table*}

\begin{table*}[htbp]
\centering
\resizebox{\textwidth}{!}{
\begin{tabular}{|l|c|c|c|c|}
\hline
\multicolumn{1}{|c|}{\multirow{2}{*}{\textbf{Dataset}}} & \multicolumn{1}{|c|}{\multirow{2}{*}{\textbf{Best Individual Prompt}}} & \multicolumn{1}{|c|}{\multirow{2}{*}{\textbf{Majority Label Voter}}} & \multicolumn{2}{c|}{\textbf{Best Ensemble Model}} \\
\centering & & & \textbf{Model} & \textbf{Bal. Acc.} \\
\hline
AggreFact-XSUM FTSOTA & 69.1 & \posval{71.1 (+2.0)} & LabelModel & \highposval{71.9 (+2.8)} \\
HaluEval Summarization & 72.7 & \posval{73.3 (+0.6)} & KNeighbors & \highposval{74.9 (+2.2)} \\
TofuEval-Summary-Level-MediaSum & 66.6 & \negval{64.3 (-2.3)} & LabelModel & \negval{66.3 (-0.3)} \\
TofuEval-Summary-Level-MeetingBank & 73.2 & \posval{74.1 (+0.9)} & LabelModel & \highposval{79.7 (+6.5)} \\
\hline
\end{tabular}
}
\caption{Showcasing the improvement in balanced accuracy achieved by ensembling compared to the top-performing individual prompts for each dataset. The number in parentheses shows the difference in balanced accuracy between the best individual prompt and the ensemble for each dataset. MajorityLabelVoter serves as our baseline ensemble, offering a simple, training-free method to combine results. Numbers in all columns are sourced from Figure~\ref{tab:ensemble_benchmarking} for ensembling results and Figure~\ref{tab:individual_prompt_performance_across_datasets} for individual prompt performance and are chosen by performance. The performance improvement in parentheses in the final column is the difference between the best ensemble model and the individual prompt performance. This represents an optimistic view of the possible the performance gain from ensembling for each dataset.}
\label{tab:comprehensive_performance_summary}
\end{table*}

\end{document}